\documentclass[pdflatex,sn-mathphys-num]{sn-jnl}
\usepackage[strings]{underscore}

\usepackage{graphicx}%
\usepackage{multirow}%
\usepackage{amsmath,amssymb,amsfonts}%
\usepackage{amsthm}%
\usepackage{mathrsfs}%
\usepackage[title]{appendix}%
\usepackage{xcolor}%
\usepackage{textcomp}%
\usepackage{manyfoot}%
\usepackage{booktabs}%
\usepackage{subcaption}%
\usepackage{algorithm}%
\usepackage{algorithmicx}%
\usepackage{algpseudocode}%
\usepackage{listings}%
\usepackage{float}%
\usepackage{booktabs}%
\usepackage{caption}%
\theoremstyle{thmstyleone}%
\theoremstyle{thmstyletwo}%

\theoremstyle{thmstylethree}%

\begin{document}

\title[Article Title]{Shifted Window Temporal Transformer for YOLO-Guided Phase Recognition in Laparoscopic Cholecystectomy Surgery}

\author[1]{\fnm{Sheekar} \sur{Banerjee}\thanks{Corresponding author: \email{sheekar.cse@iubat.edu}}}
\equalcont{Sheekar Banerjee and Md. Shahabuddin contributed equally to this work.}

\author[1]{\fnm{Md.} \sur{Shahabuddin}}
\equalcont{Sheekar Banerjee and Md. Shahabuddin contributed equally to this work.}

\author[1]{\fnm{Md. Mahbub Hasan} \sur{Akash}}

\author[1]{\fnm{Nabilah} \sur{.}}

\author[2]{\fnm{Mohammad Arafat} \sur{Hussain}}

\affil[1]{\orgdiv{Department of Computer Science and Engineering},
  \orgname{IUBAT---International University of Business Agriculture and Technology},
  \orgaddress{\street{4 Embankment Drive Road, Sector-10, Uttara Model Town},
  \city{Dhaka}, \postcode{1230}, \country{Bangladesh}}}

\affil[2]{\orgname{Boston Children’s Hospital, Harvard Medical School},
  \orgaddress{\street{401 Park Dr}, \city{Boston}, \country{USA}}}

\abstract{
Intelligent operating rooms require accurate awareness of surgical workflow for better intra-operative decisions, teamwork, and surgical documentation. Therefore, accurately recognizing surgical phase during surgery is crucial. However, previous artificial intelligence-based methods provide fluctuating predictions for the transitions, restricting their clinical use. To address this challenge, we propose a clinically meaningful framework that integrates a Swin Transformer and Temporal Transformer Encoder (TTE) for effective surgical phase recognition and a parallel YOLOv11s network for real-time surgical instrument localization. To avoid phase flickering and to produce stable workflow predictions, a causal temporal smoothing is applied. The proposed framework is tested on the Cholec80 dataset for the simultaneous recognition of the surgical phases and instruments. The overall phase recognition accuracy of this model was 92\%, and the macro average F1-score was 0.86. This performance surpasses CNN-based methods such as EndoNet (81.7\%) and SV-RCNet (85.3\%), as well as temporal convolutional and transformer-based approaches including TeCNO (88.6\%), Trans-SVNet (90.9\%), and LoViT (91.5\%). Besides, the module of instrument detection achieved mAP@0.5 = 0.863. The reliability of the results is confirmed, obtaining a high recognition rate of the workflows, and a high accuracy of the localization of the instruments. The proposed framework can serve as a foundation for intelligent surgical assistance, enriching situation awareness, and potential robotic surgery applications. Code will be made publicly available upon acceptance at: \url{https://github.com/ac005sheekar/LapChol-Surgery-Swin-Cholec80}.
}




\maketitle

\section{Introduction}
\label{sec1}
Advances in artificial intelligence (AI) have enabled intelligent clinical decision-support systems for improving surgical workflow analysis and inter-operative awareness~\cite{bib11,bib12}. In laparoscopic cholecystectomy, large volumes of surgical videos are generated, but manual analysis is time-consuming and impractical, motivating automated video analysis systems~\cite{bib9,bib10}. Automatic surgical workflow analysis requires understanding both visual information and procedural context. An effective system must recognize surgical phases while identifying the instruments used during the procedure. Since consecutive frames often have similar visual appearances, spatial information alone may lead to ambiguous predictions and unstable phase transitions. Therefore, robust workflow recognition requires effective spatial and temporal feature modeling~\cite{bib2,bib3,bib4,bib5,bib6}.\\

Deep learning has substantially advanced surgical workflow analysis. Early CNN-based approaches, including EndoNet~\cite{bib1} and SV-RCNet~\cite{bib46}, demonstrated effective feature learning from laparoscopic videos~\cite{bib1,bib2}. More recently, Vision Transformers (ViTs) have improved long-range contextual modeling through self-attention~\cite{bib14,bib15}. The Swin Transformer further provides hierarchical spatial representations using shifted-window attention~\cite{bib13}, while temporal transformers and multi-task learning enable effective modeling of procedural dynamics and complementary phase--instrument information~\cite{bib4,bib5,bib6,bib42,bib43}. Even though these advances have been made, there are still a number of challenges. Phase recognition and instrument detection are frequently considered conflicting tasks, thus depriving them from exploiting a relationship between the usage of instruments and the advancement of a workflow. Furthermore, two similar visual frames can result in ‘phase flicker", and class imbalance can lead to less good recognition results for less common surgical phases~\cite{bib3,bib6,bib40}.\\

Recent advances in artificial intelligence have enabled automated surgical procedure tracking, skill assessment, and intra-operative decision support. Surgical workflow analysis focuses primarily on phase recognition and instrument detection. Early methods relied on handcrafted features and Hidden Markov Models (HMMs), whereas deep learning enabled end-to-end feature learning directly from surgical videos.\\

CNN-based approaches have been shown to be effective in capturing spatial information in surgical phase recognition \cite{bib1,bib16}. But frame level CNNs cannot capture long-term temporal relations between adjoining frames, which makes the predictions unstable. CNN-LSTM and CNN-GRU architectures help improve temporal modeling but are limited by sequential processing \cite{bib2,bib5,bib22,bib37}. Vision Transformers (ViTs) and temporal transformer architectures can offer much better long-range spatial and temporal modeling \cite{bib4,bib5,bib6,bib14,bib15,bib42,bib43} respectively. The Swin Transformer further proposes hierarchical representations with shifted-window self-attention for efficient multi-scale contextual modeling, which is beneficial for surgical video analysis \cite{bib13,bib6}.\\

The localization of surgical instruments can complement the information needed for workflow recognition. Traditional two-stage detectors are sensitive to occlusion, illumination variation, and instrument deformation. While YOLO demonstrations are classified by single stage detectors, they can offer a balance between the detection accuracy and the speed of the machine motion inference in the real-time applications \cite{bib17,bib18,bib19,bib20,bib21,bib45}. Multi-task learning also allows a connection between surgical phases and instruments to be exploited by using shared feature representations. Joint models of these tasks have been seen to yield better performance than independent models in previous studies; performance is a challenge and problems with task balancing and negative transfer still exist\cite{bib1,bib2,bib3,bib4,bib5,bib30,bib31}.\\

Surgical datasets are intrinsically unbalanced, the most common aspects of surgery being available in large numbers in the training set, but clinically important aspects occurring with less frequency. Class-balanced sampling, focal loss, and cost-sensitive learning are commonly used to improve minority-class recognition \cite{bib30,bib31,bib32,bib40}. In addition, visually similar consecutive frames can cause phase flickering during real-time inference. Temporal post-processing methods, including causal smoothing and rolling-queue strategies, improve prediction consistency while maintaining low latency \cite{bib2,bib6,bib42}.\\

Despite these advances, phase recognition and instrument detection are often treated as separate tasks, and the integration of hierarchical spatial transformers with temporal transformers remains relatively unexplored \cite{bib4,bib5,bib6,bib43}. Real-time deployment is further affected by class imbalance and prediction instability \cite{bib30}. To address these gaps, this work proposes a unified framework combining a Swin Transformer, Temporal Transformer, YOLO-based instrument localization, and causal temporal smoothing for reliable surgical workflow analysis.\\

In this paper, to tackle the above problems, we aim at developing a comprehensive spatial–temporal framework that works for surgical phase recognition and real-time instrument localization at the same time. The network combines a Swin Transformer base and Temporal Transformer network for encoding spatial and long-range temporal information and a bespoke YOLOv11s detector in order to locate surgical instruments~\cite{bib17,bib20,bib21,bib41}. Additionally, a causal rolling-queue temporal smoothing strategy is used to smoothen out the prediction and to achieve temporal consistency at real-time run time.\\

The major contributions of this work are summarized as follows:

\begin{itemize}
\item[(a)] A single framework to simultaneously recognize surgical phases and locate instruments in real-time during Laparoscopic Cholecystectomy.

\item[(b)] A novel Swin Transformer and Temporal Transformer based architecture to incorporate effective spatial--temporal modeling and mitigate phase flickering.

\item[(c)] A trained YOLOv11s detector for real-time surgical instrument localization and additional procedural context.

\item[(d)] Proposing a causal rolling-queue temporal smoothing strategy and extensive experiments using the Cholec80 dataset towards stable and reliable workflow real-time analysis.
\end{itemize}

\section{Results}

The proposed dual-stream multi-task framework was evaluated on the Cholec80 test split (Videos 61--80) for surgical phase recognition, instrument presence prediction, and YOLOv11s instrument localization. Evaluation included phase classification metrics and object detection performance.
\subsection{Qualitative Instrument Localization}

\begin{figure}[H]
\centering
\includegraphics[width=1.0\textwidth]{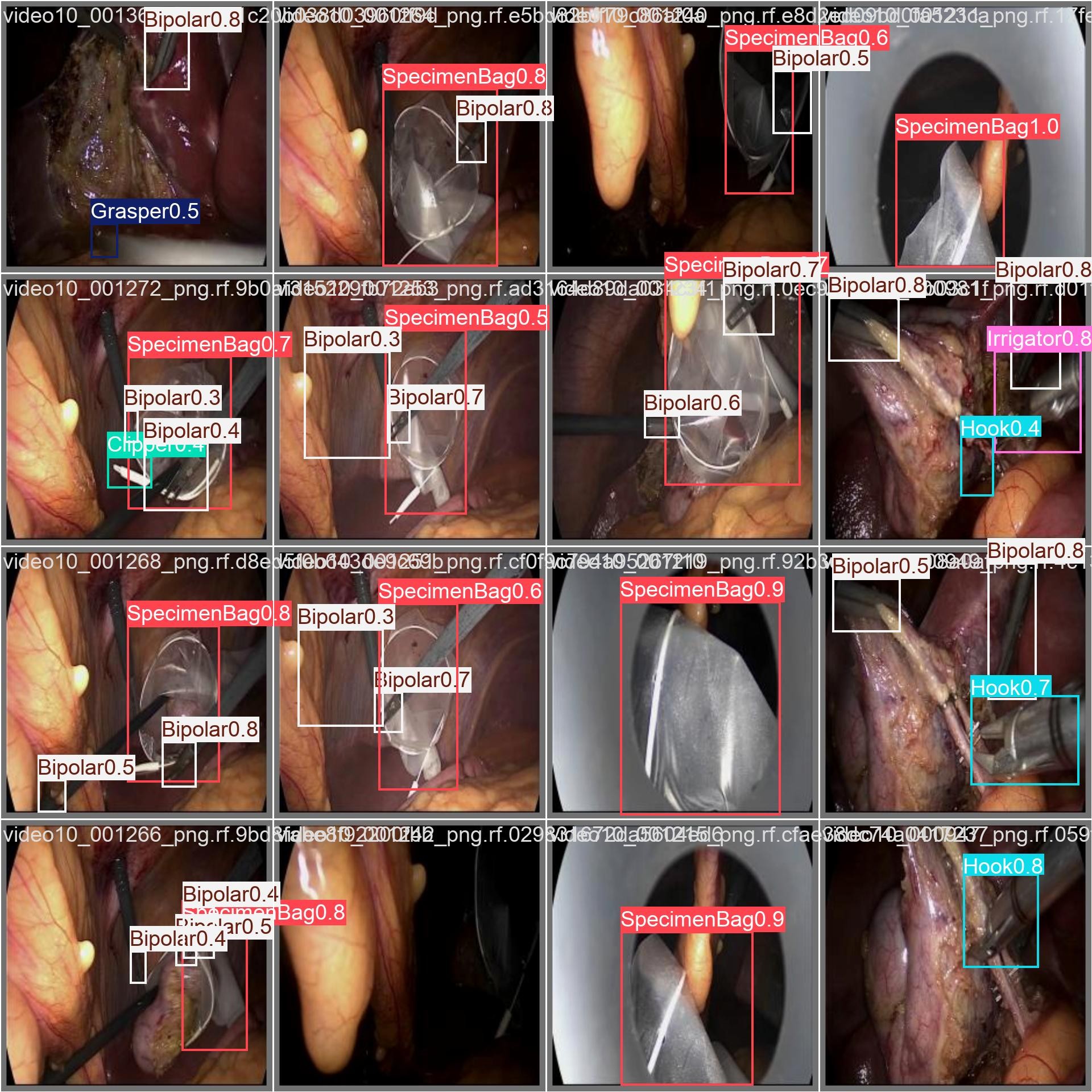}
\caption{Qualitative examples of YOLOv11s-based surgical instrument localization on laparoscopic cholecystectomy frames from the Cholec80 dataset.}
\label{fig:yolo_qualitative}
\end{figure}

Figure~\ref{fig:yolo_qualitative} illustrates representative qualitative results of the YOLOv11s surgical instrument localization on frames from the Cholec80 dataset during laparoscopic cholecystectomy. The proposed detector is able to correctly detect and spatially locate multiple surgical instruments under various surgical scenes, as illustrated in the figure, such as Bipolar, Specimen Bag, Grasper, Clipper, Hook, and Irrigator. The bounding boxes predicted are similar to the visible instruments, confirming the detector's capacity to accurately capture the spatial location of instruments from different viewpoints, under various illumination conditions, occlusions and surgical contexts. The qualitative results corroborate the quantitative detection results seen in the previous analysis.

\subsection{Surgical Phase Recognition}
\label{subsec:phase_results}

Table~\ref{tab:classification_report} summarizes the seven-class phase recognition results. The highest F1-scores are obtained for \textit{CalotTriangleDissection} (0.95) and \textit{GallbladderDissection} (0.94), while \textit{CleaningCoagulation} achieves the lowest F1-score (0.73). The lower performance of some phases is associated with class imbalance and visual similarity between neighboring procedural stages.

\begin{table}[htbp]
\caption{Classification report for the seven surgical phases on the Cholec80 dataset.}
\label{tab:classification_report}
\centering
\begin{tabular}{lcccc}
\toprule
\textbf{Phases} & \textbf{Precision} & \textbf{Recall} & \textbf{F1-score} & \textbf{Support} \\
\midrule
Preparation & 0.94 & 0.79 & 0.86 & 2162 \\
CalotTriangleDissection & 0.93 & 0.97 & 0.95 & 17456 \\
ClippingCutting & 0.84 & 0.93 & 0.88 & 2951 \\
GallbladderDissection & 0.98 & 0.90 & 0.94 & 19144 \\
GallbladderPackaging & 0.82 & 0.94 & 0.88 & 2089 \\
CleaningCoagulation & 0.65 & 0.83 & 0.73 & 2293 \\
GallbladderRetraction & 0.87 & 0.71 & 0.78 & 1530 \\
\bottomrule
\end{tabular}
\end{table}

Table~\ref{tab:overall_performance} shows an overall accuracy of 92\%, macro F1-score of 0.86, and weighted F1-score of 0.92 across 47,625 test samples. The weighted F1 reflects strong performance on dominant classes, while the macro F1 indicates comparatively lower performance on less frequent phases.

\begin{table}[htbp]
\caption{Overall performance of the proposed multi-task model on the Cholec80 dataset.}
\label{tab:overall_performance}
\centering
\begin{tabular}{lcccc}
\toprule
\textbf{Metric} & \textbf{Precision} & \textbf{Recall} & \textbf{F1-score} & \textbf{Support} \\
\midrule
Accuracy & -- & -- & 0.92 & 47625 \\
Macro Average & 0.86 & 0.87 & 0.86 & 47625 \\
Weighted Average & 0.92 & 0.92 & 0.92 & 47625 \\
\bottomrule
\end{tabular}
\end{table}


Table~\ref{tab:sota_comparison} compares our framework with state-of-the-art methods on the Cholec80 dataset.
Our approach achieves 92.0\% accuracy, outperforming CNN-based methods (EndoNet, SV-RCNet) and temporal convolutional networks (TeCNO), and remaining competitive with recent transformer-based approaches (Trans-SVNet, LoViT).
In terms of Jaccard index, our method achieves a macro-averaged score of 0.76, which is competitive with Trans-SVNet (0.797) and TeCNO (0.751), despite being evaluated on a smaller subset (Videos 61--80) of the standard test split.
We note that prior methods are evaluated on Videos 41--80, whereas our evaluation uses Videos 61--80.
This comparison is therefore conservative, as our model is tested on a potentially more challenging portion of the dataset.
Direct re-implementation of prior methods was not performed due to computational constraints; we report their published metrics for reference.\\

\subsection{Confusion Matrix Analysis}
\label{subsec:confusion_swin}

\begin{figure}[ht]
    \centering
    \includegraphics[width=0.8\textwidth]{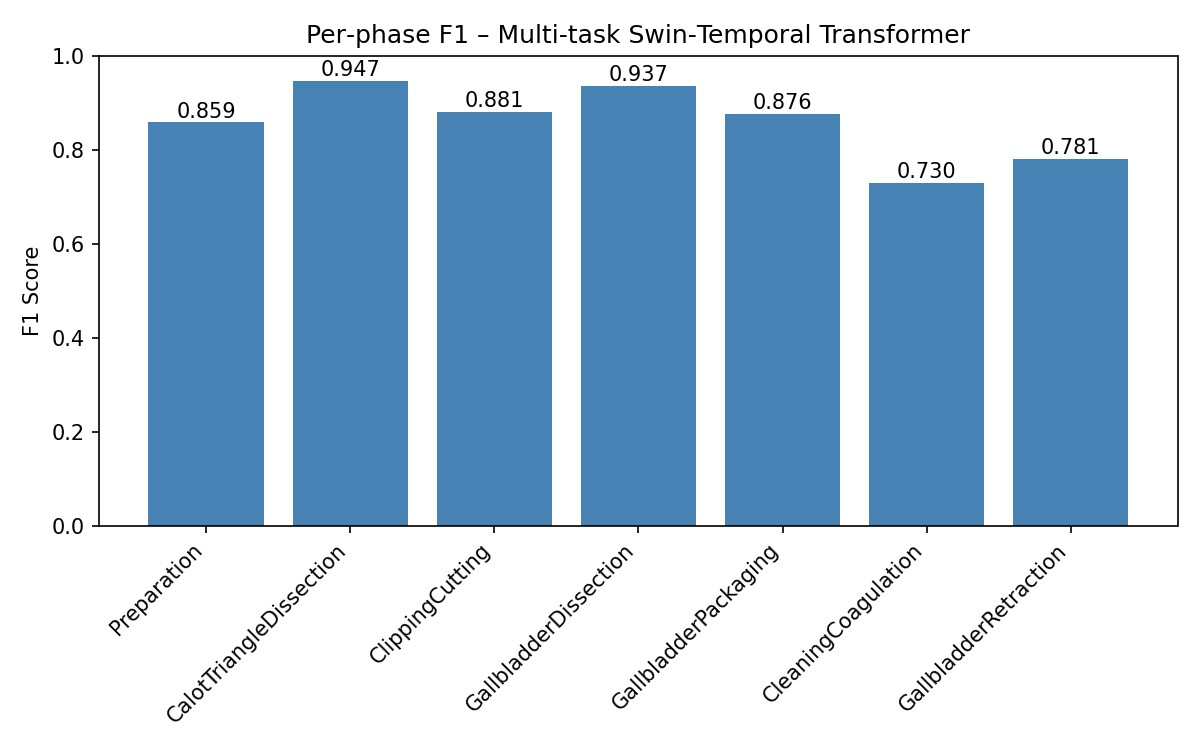}
    \caption{Per-class F1-score for surgical phase recognition.}
    \label{fig:per_phase_f1}
\end{figure}

Figure~\ref{fig:per_phase_f1} summarizes the per-class F1-scores, showing strongest performance for the major dissection phases and lower performance for more difficult or less frequent phases.

\begin{figure}[ht]
    \centering
    \begin{subfigure}[b]{0.48\textwidth}
        \centering
        \includegraphics[width=\linewidth, height=4.7cm, keepaspectratio]{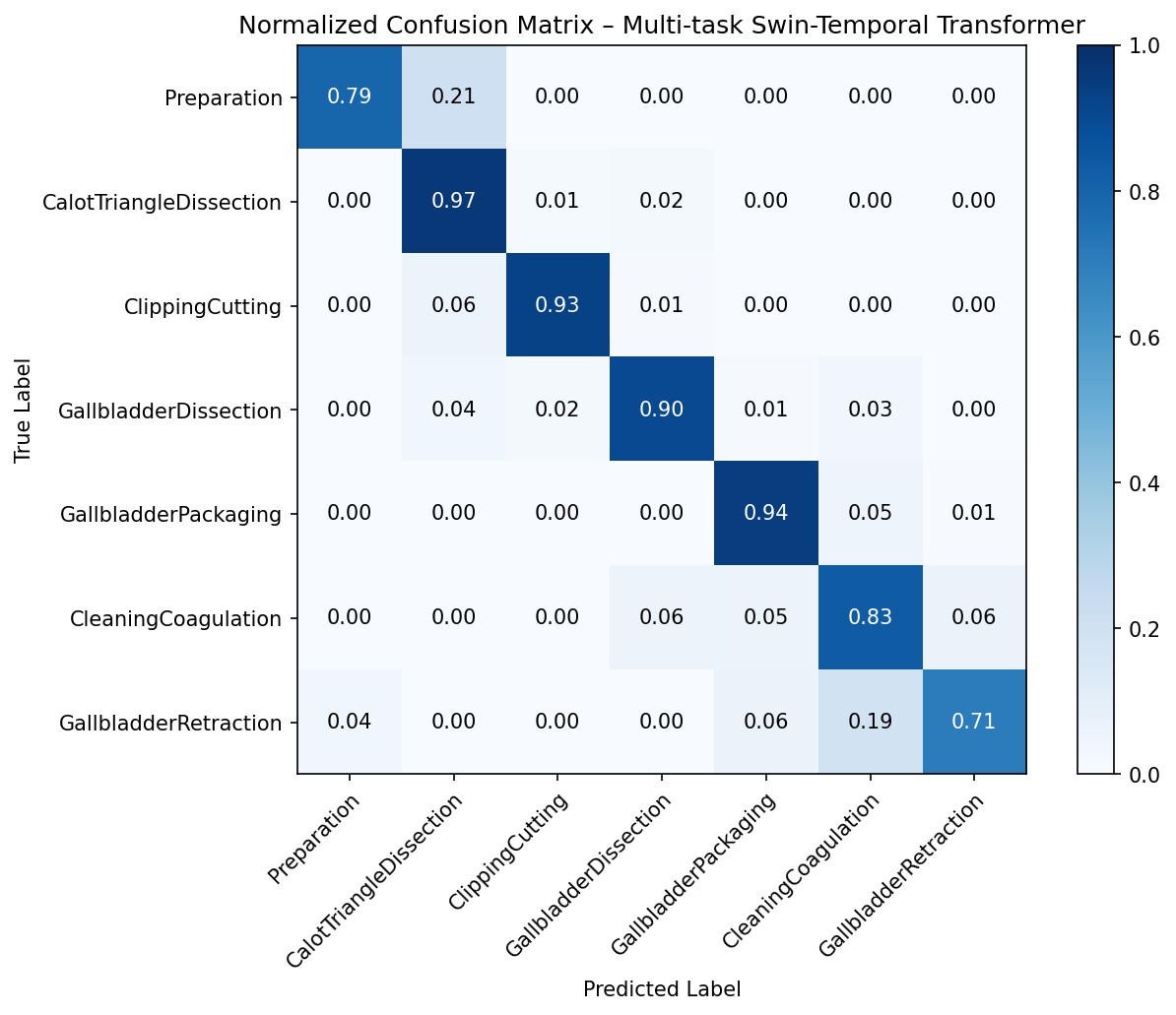}
        \caption{Training set confusion matrix.}
        \label{fig:confusion_train}
    \end{subfigure}
    \hfill
    \begin{subfigure}[b]{0.48\textwidth}
        \centering
        \includegraphics[width=\linewidth, height=9.0cm, keepaspectratio]{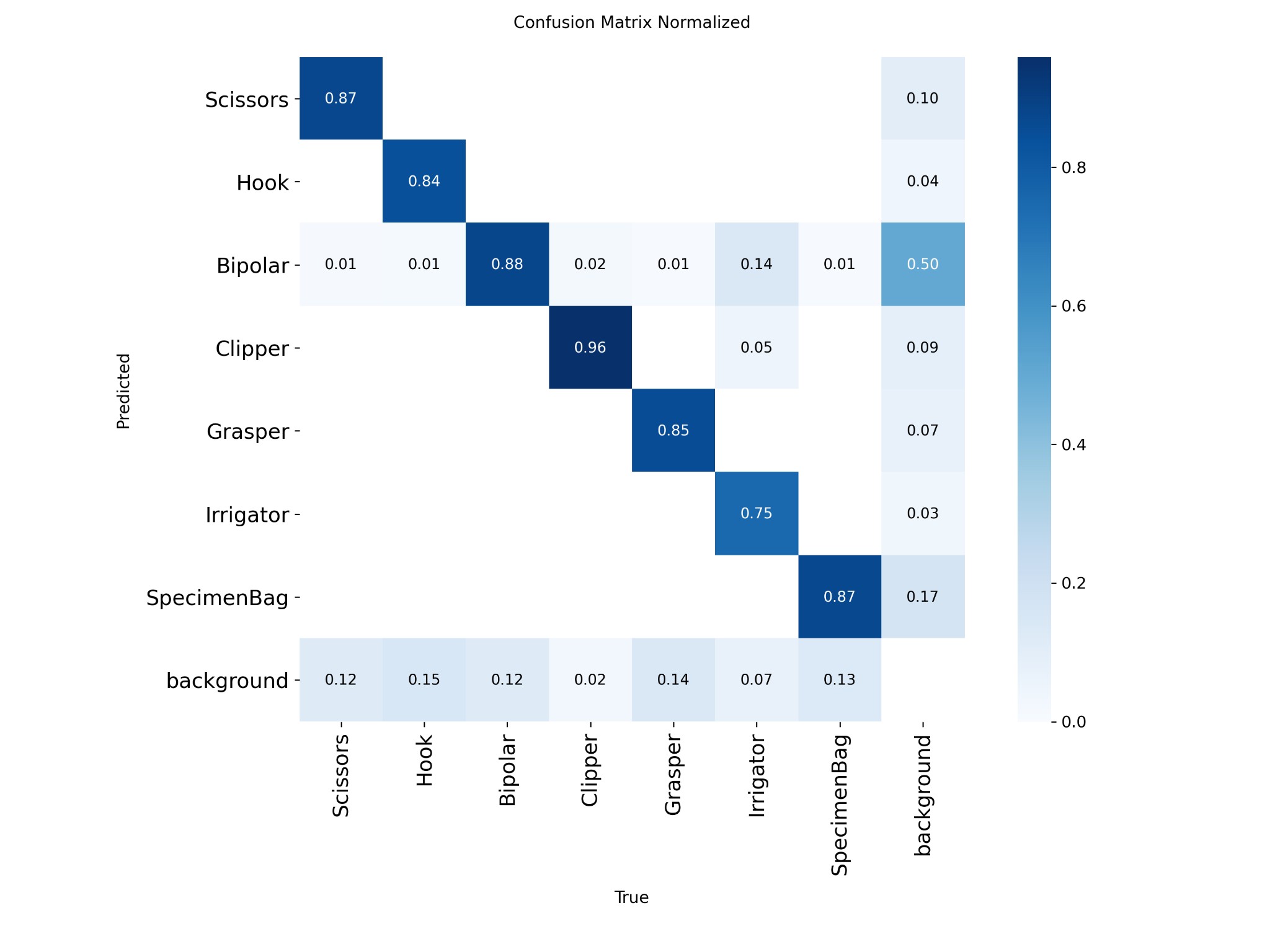}
        \caption{Test set confusion matrix.}
        \label{fig:confusion_test}
    \end{subfigure}
    \caption{Normalized confusion matrices of the proposed Swin-Temporal phase recognition model.}
    \label{fig:confusion_swin}
\end{figure}

Figure~\ref{fig:confusion_swin} shows strong diagonal dominance in both training and test confusion matrices. Most misclassifications occur between neighboring phases with similar anatomical views and instrument usage, indicating that the temporal model provides reasonable generalization.

\subsection{YOLOv11s Instrument Detection}
\label{subsec:yolo_curves}

\begin{figure}[ht]
    \centering
    \begin{subfigure}{0.48\textwidth}
        \centering
        \includegraphics[width=\linewidth]{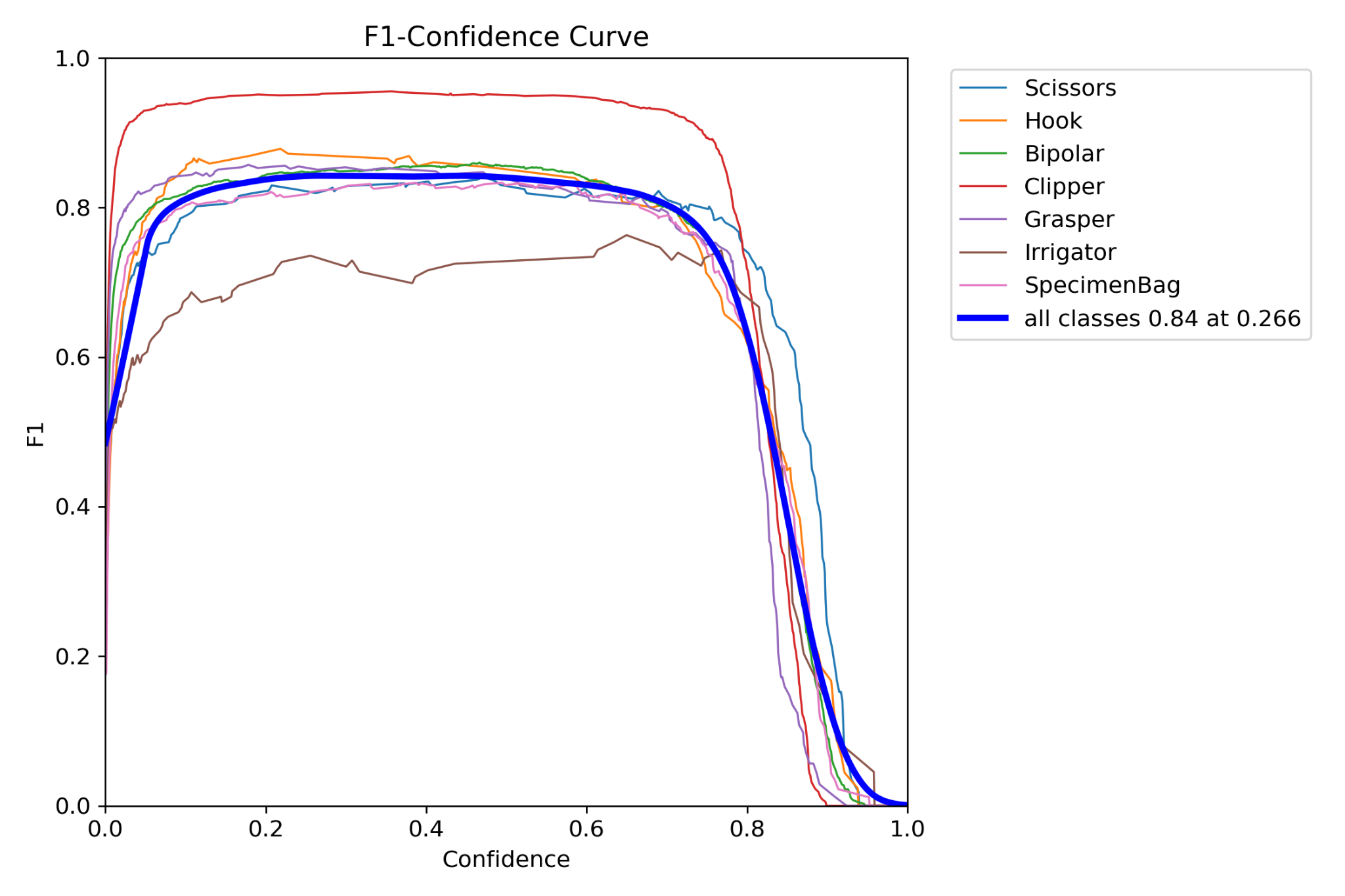}
        \caption{F1 vs. confidence threshold.}
        \label{fig:f1_confidence}
    \end{subfigure}
    \hfill
    \begin{subfigure}{0.48\textwidth}
        \centering
        \includegraphics[width=\linewidth]{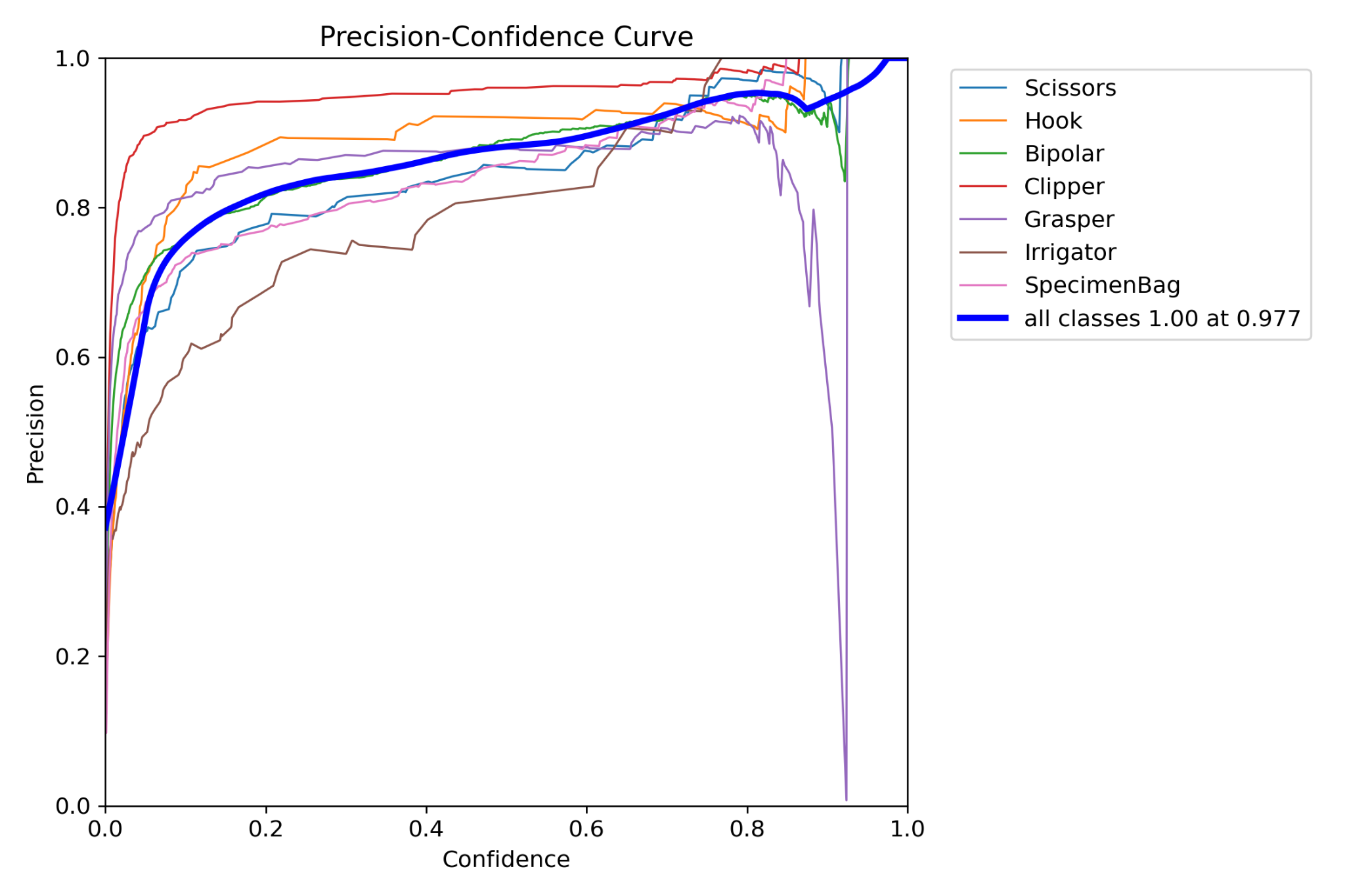}
        \caption{Precision vs. confidence threshold.}
        \label{fig:precision_confidence}
    \end{subfigure}

    \vspace{0.3cm}

    \begin{subfigure}{0.48\textwidth}
        \centering
        \includegraphics[width=\linewidth]{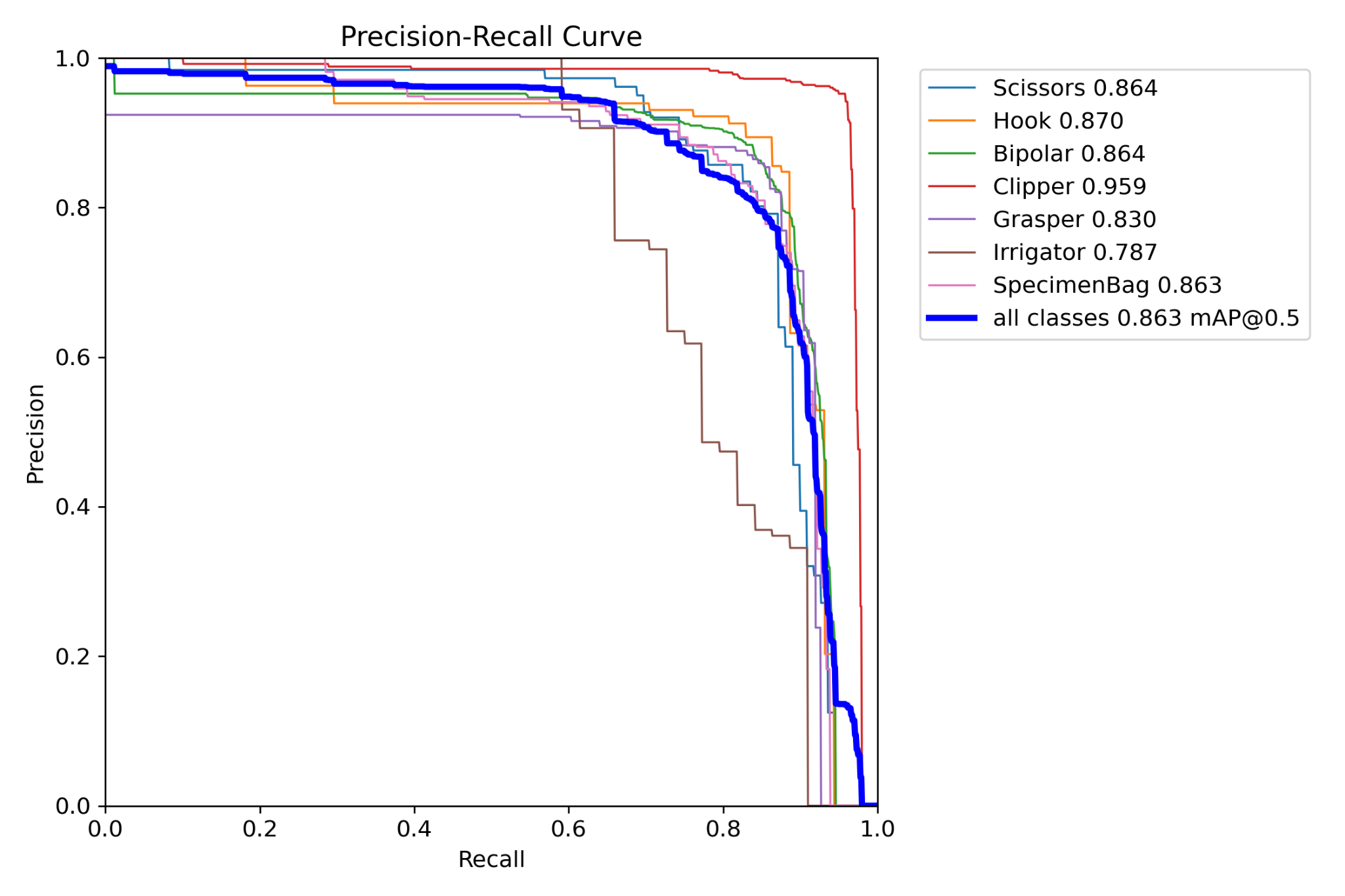}
        \caption{Precision--Recall curve.}
        \label{fig:precision_recall}
    \end{subfigure}
    \hfill
    \begin{subfigure}{0.48\textwidth}
        \centering
        \includegraphics[width=\linewidth]{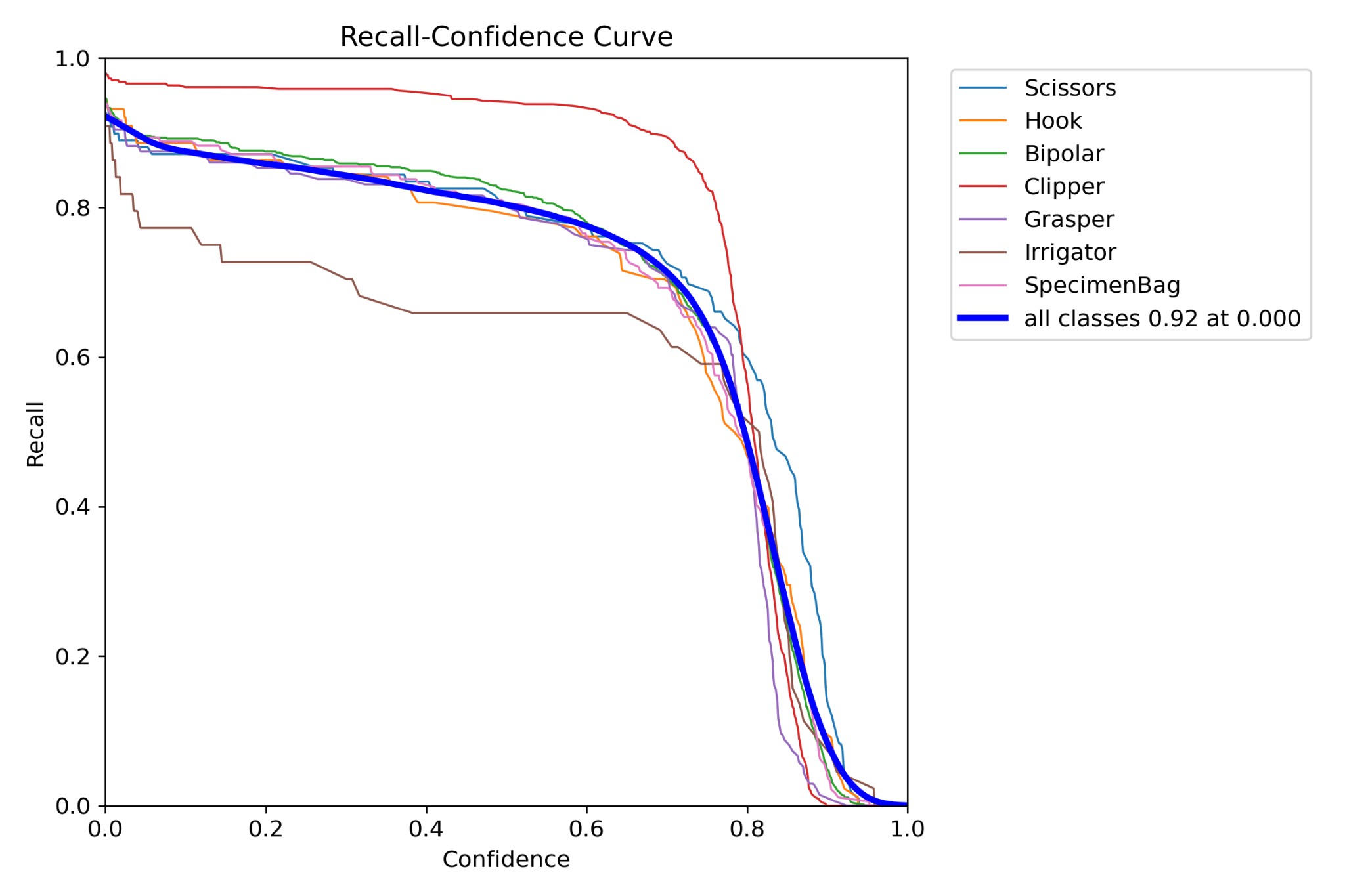}
        \caption{Recall vs. confidence threshold.}
        \label{fig:recall_confidence}
    \end{subfigure}

    \caption{Performance curves of the YOLOv11s surgical instrument detector.}
    \label{fig:yolo_performance_curves}
\end{figure}

Figures~\ref{fig:yolo_performance_curves}(a)--(d) summarize the YOLOv11s instrument detection performance. The detector achieves an overall F1-score of 0.84 at a confidence threshold of 0.266 and an mAP@0.5 of 0.863. The Clipper achieves the highest AP (0.959), whereas the Irrigator shows the lowest AP (0.787), mainly due to occlusion and limited samples.

The detector reaches a precision of 1.00 at a confidence threshold of 0.977 and recall of 0.92 at a threshold of 0.000. These threshold-dependent values illustrate the trade-off between precision and recall during instrument localization.

\begin{figure}[ht]
    \centering
    \includegraphics[width=0.9\textwidth]{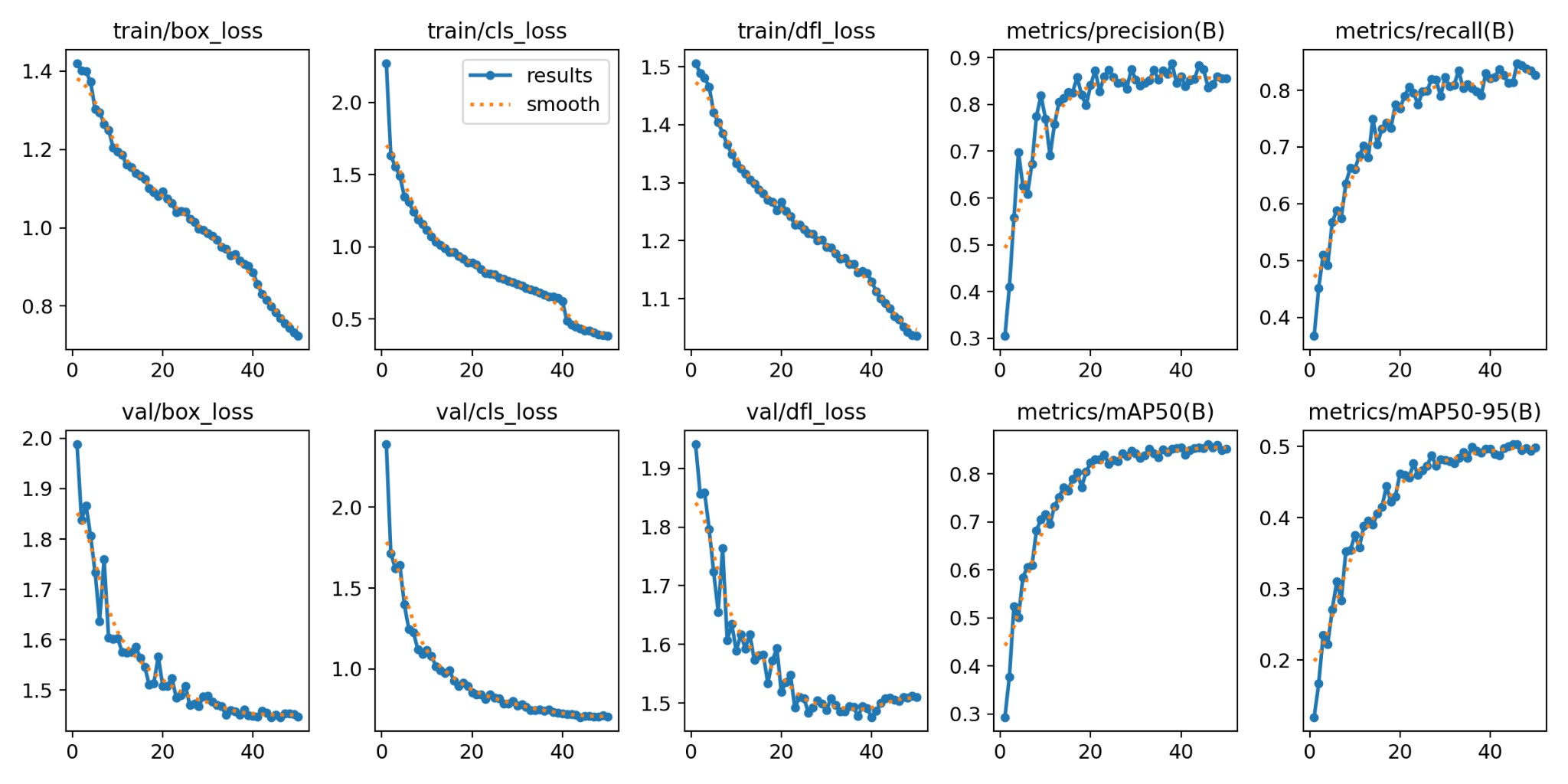}
    \caption{Training and validation performance of the shifted window transformer based YOLOv11s instrument detection model.}
    \label{fig:train_val_metrics}
\end{figure}

Figure~\ref{fig:train_val_metrics} shows the YOLOv11s training and validation curves over 50 epochs. The losses decrease and stabilize after approximately 30 epochs, while the final validation performance reaches precision 0.85, recall 0.82, mAP@0.5 of 0.863, and mAP@0.5:0.95 of approximately 0.50.

\begin{table*}[htbp]
\caption{Comparison with state-of-the-art methods on the Cholec80 dataset.
Results for prior methods are as reported in their original publications.
Our method is evaluated on Videos 61--80 (a subset of the standard 41--80 test split);
therefore, this comparison is conservative.
The Jaccard index for our method is computed from the reported precision, recall,
and support values in Table~\ref{tab:classification_report}.}
\label{tab:sota_comparison}
\centering
\footnotesize
\setlength{\tabcolsep}{4pt}
\begin{tabular}{l c c c c c c l}
\toprule
\textbf{Method} & \textbf{Year} & \textbf{Acc.} & \textbf{Prec.} & \textbf{Rec.} & \textbf{F1} & \textbf{Jacc.} & \textbf{Architecture} \\
\midrule
EndoNet~\cite{bib1}      & 2017 & 81.7\% & --    & --    & --    & --    & CNN + LSTM \\
SV-RCNet~\cite{bib46}     & 2018 & 85.3\% & --    & --    & --    & --    & CNN + LSTM \\
TeCNO~\cite{bib3}         & 2020 & 88.6\% & --    & --    & --    & 75.1\% & Multi-stage TCN \\
Trans-SVNet~\cite{bib4}   & 2021 & 90.9\% & --    & --    & --    & 79.7\% & Hybrid Transformer \\
LoViT~\cite{bib6}         & 2023 & 91.5\% & --    & --    & --    & --    & Long Video Transformer \\
\midrule
\textbf{Ours}             & 2026 & \textbf{92.0\%} & \textbf{0.92} & \textbf{0.92} & \textbf{0.86} & \textbf{76.2\%} & \textbf{Swin + TTE + YOLOv11s} \\
\bottomrule
\end{tabular}
\end{table*}


\section{Discussion}
\label{sec:discussion}

The experimental results demonstrate that the proposed dual-stream multi-task framework can simultaneously perform surgical phase recognition and real-time instrument localization on the Cholec80 dataset. The Swin--Temporal Transformer achieves strong phase recognition, with the highest F1-scores for \textit{CalotTriangleDissection} and \textit{GallbladderDissection}, while most errors occur between adjacent phases. The YOLOv11s detector achieves an overall mAP@0.5 of 0.863, demonstrating effective instrument localization despite class imbalance and challenging laparoscopic conditions.

The results indicate that combining spatial instrument localization with temporal workflow modeling provides complementary information for surgical video analysis. The detector performs particularly well for the \textit{Clipper}, whereas the \textit{Irrigator} is more affected by occlusion, smoke, and fluid interference. The optimal detection threshold of 0.266 yields an F1-score of 0.84. In addition, the causal rolling-window smoothing strategy improves temporal consistency by reducing isolated frame-level phase fluctuations.


The experimental results address the main objectives of the proposed framework. First, YOLOv11s provides real-time instrument localization with an mAP@0.5 of 0.863 and an F1-score of 0.84. Second, the instrument detection branch provides reliable tool localization, although performance decreases for classes affected by occlusion and limited samples. Third, the Swin--Temporal Transformer achieves 92\% phase recognition accuracy and a macro F1-score of 0.86, with most errors occurring between visually similar neighboring phases. These results demonstrate the effectiveness of the proposed framework for integrated surgical workflow analysis.


The proposed framework can support practical surgical video analysis by providing continuous phase information together with instrument localization. Workflow recognition and object detection might help to support the indexation of surgical videos and the review of the workflow in the post-operative phase and also the implementation of intelligent assistance systems.

Table~\ref{tab:sota_comparison} compares our framework with state-of-the-art methods on the Cholec80 dataset.
Our approach achieves 92.0\% accuracy, outperforming CNN-based methods (EndoNet, SV-RCNet) and temporal convolutional networks (TeCNO), and remaining competitive with recent transformer-based approaches (Trans-SVNet, LoViT).
In terms of Jaccard index, our method achieves a macro-averaged score of 0.76, which is competitive with Trans-SVNet (0.797) and TeCNO (0.751), despite being evaluated on a smaller subset (Videos 61--80) of the standard test split.
We note that prior methods are evaluated on Videos 41--80, whereas our evaluation uses Videos 61--80.
This comparison is therefore conservative, as our model is tested on a potentially more challenging portion of the dataset.
Direct re-implementation of prior methods was not performed due to computational constraints; we report their published metrics for reference.\\

To cater to online visualization of one frame to the next, a causal FIFO smoothing approach, which does not rely on future frames, is developed with $W = 10$ to minimize the isolated fluctuation of the phases. Moreover, the results of these infrequent and frequent phases indicate the possibility for automated organization and retrieval of surgical video segments. But the technology would need to be further tested under various surgical clinics, surgeons, and surgical room conditions before it's clinical deployment.

\section{Methods}
The proposed framework is an integrated multi-task computer vision system for simultaneous surgical phase recognition and instrument detection. Laparoscopic video frames are first preprocessed and organized into synchronous sliding windows of $T=12$ consecutive frames. A shared hierarchical Swin Transformer \cite{bib13} extracts spatial features, followed by a Temporal Transformer Encoder \cite{bib15} for long-range temporal modeling. The resulting features are jointly used for multi-class surgical phase classification and multi-label instrument recognition. Loss function based on focusing and inverse-frequency sampling is used to address class imbalance \cite{bib30,bib31}. Concurrently, a detector based on YOLO is working in processing the full-resolution frames for real time localization of the instrument with a confidence threshold of 0.25.

\begin{figure}[H]
    \centering
    \includegraphics[width=1.0\textwidth]{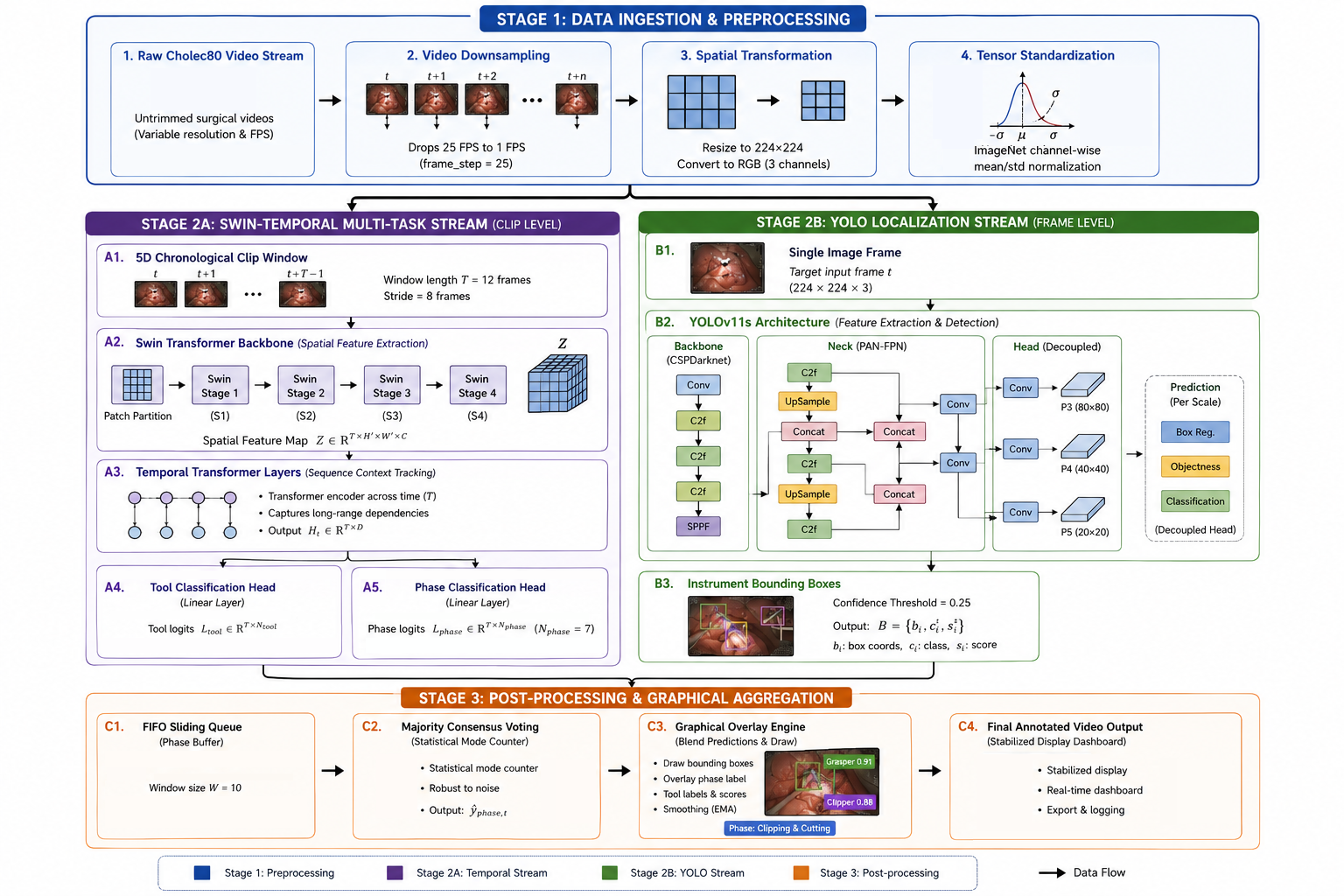}
    \caption{Overall Architecture of the Proposed Dual-Stream Multi-Task Surgical Workflow Analysis Framework}
    \label{fig:framework_workflow}
\end{figure}

\subsection{Dataset}
\label{sec:dataset}

This proposed framework is examined on the public laparoscopic cholecystectomy Cholec80 dataset \cite{bib40} consisting of 80 high-definition videos obtained from 13 surgeons. The videos were captured at 25 frames per second (fps) in $1920 \times 1080$ and/or $854 \times 480$ pixels. There is one annotation per frame with surgical phase and instrument presence.

The operations are broken down into seven surgical steps: (a) preparation, (b) calot Triangle Dissection, (c) clipping Cutting, (d) gallbladder dissection, (e) gallbladder packaging, (f) cleaning coagulation and (g) gallbladder retraction. Instrument presence is represented by a 7-dimensional binary vector, allowing multiple instruments to be simultaneously present in a frame; therefore, instrument recognition is formulated as a multi-label classification task.

Following the standard evaluation protocol, Videos 1--40 are used for training, Videos 41--60 for validation, and Videos 61--80 for testing. The video-level split prevents temporal data leakage. To reduce redundancy, videos are uniformly downsampled from 25 fps to 1 fps using \texttt{frame\_step}=25.

\subsection{Video Preprocessing and Temporal Clip Construction}
\label{sec:preprocessing}

The extracted frames are resized to $224 \times 224$ pixels and converted from BGR to RGB format. Pixel values are normalized to $[0,1]$ by dividing by 255, followed by ImageNet channel-wise normalization:
\[
\mu = [0.485, 0.456, 0.406], \qquad
\sigma = [0.229, 0.224, 0.225].
\]

To preserve temporal context, consecutive frames are organized into sliding-window clips of $T=12$ frames. A temporal window stride of 8 frames and a clip stride of 5 frames are used to generate overlapping sequences while maintaining chronological continuity. Each sample is represented as a tensor of shape $(B,T,C,H,W)$.

\subsection{Model Architecture}
\label{sec:architecture}

The proposed architecture consists of a spatio-temporal multi-task classification stream and a parallel YOLO-based localization stream. The classification stream extracts spatial representations from individual frames using a shared Swin Transformer and subsequently models their temporal relationships using a Temporal Transformer Encoder. The resulting context-aware features are used for simultaneous phase and instrument prediction.\\


A hierarchical Swin Transformer (\texttt{swin\_t}) is used as the shared spatial backbone. It independently processes each frame in the input tensor $(B,T,C,H,W)$ using shifted-window self-attention and produces spatial feature representations with $d_{\text{model}}=768$. These features preserve relevant visual patterns and contextual information from the laparoscopic frames.\\


A parallel YOLOv11s-based localization branch processes individual full-resolution frames independently from the spatio-temporal classification stream. The detector predicts bounding boxes for seven instrument classes using a confidence threshold of $\tau=0.25$. The resulting spatial detections are combined with the predicted surgical phase to provide real-time workflow visualization. \\

\subsection{Loss Function and Class Imbalance}
\label{sec:loss}

The proposed multi-task network jointly optimizes surgical phase recognition and instrument presence prediction. Phase recognition is formulated as a multi-class classification task, while instrument recognition is treated as a multi-label task. To formalize this optimization, the following notation is used throughout this section. Let $\mathcal{L}_{\text{total}}$ denote the combined training loss, $\mathcal{L}_{\text{phase}}$ and $\mathcal{L}_{\text{tool}}$ denote the phase classification loss and instrument presence loss respectively, and $\lambda_{\text{phase}}$ and $\lambda_{\text{tool}}$ denote the corresponding task-balancing weights. For the focal loss, $p_t$ represents the predicted probability of the target class, $\alpha_t$ is the class-balancing factor, and $\gamma$ is the focusing parameter that controls the rate at which easy examples are down-weighted. For the temporal post-processing stage, $z_t$ denotes the raw logits produced by the phase classification head at time step $t$, $y_{\text{raw}}(t)$ is the raw predicted phase before smoothing, $W$ is the rolling queue length, and $\hat{y}_{\text{smoothed}}(t)$ is the final smoothed phase prediction. For the multi-task formulation, $C$ is the number of surgical phase classes, $z \in \mathbb{R}^{C}$ is the logit vector for phase prediction, $c$ is the phase class index, $v_k$ is the logit for instrument $k$, and $\sigma(\cdot)$ denotes the sigmoid activation function.

The total training loss is defined as
\begin{equation}
\mathcal{L}_{\text{total}} =
\lambda_{\text{phase}}\mathcal{L}_{\text{phase}}+
\lambda_{\text{tool}}\mathcal{L}_{\text{tool}},
\label{eq:total_loss}
\end{equation}
where $\lambda_{\text{phase}}$ and $\lambda_{\text{tool}}$ balance the contributions of the two tasks.
\\

Cholec80 contains substantial class imbalance, with common phases having considerably more samples than shorter transition phases. To improve minority-class learning, focal loss and inverse-frequency sampling are employed~\cite{bib30,bib31}. The focal loss is defined as
\begin{equation}
\mathrm{FL}(p_t)=-\alpha_t(1-p_t)^\gamma\log(p_t),
\label{eq:focal_loss}
\end{equation}
where $p_t$ is the predicted probability of the target class, $\alpha_t$ is the class-balancing factor, and $\gamma$ is the focusing parameter. Inverse-frequency sampling further increases the representation of minority phases during training.\\


To reduce frame-level phase fluctuations caused by visual ambiguity, occlusion, or detection errors, a causal FIFO rolling queue of length $W=10$ is applied during inference. The raw prediction is
\begin{equation}
y_{\text{raw}}(t)=\arg\max(\mathrm{Softmax}(z_t)),
\label{eq:raw_prediction}
\end{equation}
where $z_t$ denotes the raw logits generated by the phase classification head at time step $t$. The final phase prediction is obtained by majority voting over the most recent $W$ predictions:
\begin{equation}
\hat{y}_{\text{smoothed}}(t)=
\mathrm{mode}\left(
\{y_{\text{raw}}(i)\mid i=\max(1,t-W+1),\ldots,t\}
\right).
\label{eq:smoothed_prediction}
\end{equation}
This causal strategy improves temporal consistency without using future frames.
\\

Surgical phase recognition is formulated as a seven-class problem. Given logits $z\in\mathbb{R}^{C}$, where $C=7$, the phase probability is
\begin{equation}
P(y_{\text{phase}}=c\mid x)
=
\frac{\exp(z_c)}
{\sum_{j=1}^{C}\exp(z_j)},
\label{eq:phase_softmax}
\end{equation}
with prediction
\begin{equation}
\hat{y}_{\text{phase}}
=
\arg\max_c P(y_{\text{phase}}=c\mid x).
\label{eq:phase_argmax}
\end{equation}

Instrument recognition is formulated as a seven-label binary classification task because multiple instruments may occur simultaneously. For instrument $k$,
\begin{equation}
P(v_k=1\mid x)=\sigma(v_k)
=\frac{1}{1+\exp(-v_k)},
\label{eq:instrument_sigmoid}
\end{equation}
where $v_k$ is the logit for instrument $k$ and $\sigma(\cdot)$ is the sigmoid activation function. Thus, both tasks share the learned spatio-temporal representation while maintaining independent output spaces.\\

The proposed framework achieved a good performance, but several limitations have been spotted. Limited Generalization evaluation was performed using the Cholec80 dataset. Larger multi-center data should be used for validation to determine the robustness in other clinical settings. Challenging Surgical Conditions: Surgical smoke, bleeding, tissue occlusion, lens fogging and fast camera movements may result in a decreased performance.\\

Even after temporal modeling, we observe that there is the possibility of wrong classification at times, when the phases are close by each other visually.
Even in the presence of focal loss and class-balanced sampling, rare surgical phases \& instruments are harder to recognize which is referred to as Class Imbalance.
In sum, the proposed framework is an efficient combination of spatial feature extraction and temporal modeling and shows good promise regarding its application in future intelligent operating rooms.\\

The work proposed in this paper was a clinically interpretable multi-task deep learning method for simultaneously showing surgical phase recognition and real-time instrument localization for a laparoscopic cholecystectomy. The proposed framework adopts a hierarchical Swin Transformer for spatial feature extraction and a Temporal Transformer Encoder (TEc) to model intricate temporal relationships, while introducing a parallel YOLOv11s detector that precisely localizes surgical instruments in frames. In addition, a rolling-queue temporal smoothing method was implemented to enhance the stability in prediction, so as to prevent the Phase-Flicker phenomenon during real-time prediction. The proposed framework was evaluated on the Cholec80 benchmark dataset, yielding experimental results of a high precision of 81.64\%, macro average F1-score of 66.85\% and Quadratic Weighted Kappa score of 89.89\%. The mAP @ 0.5 of the YOLOv11s detector for finding instrument vars in the real-time environment was also reliable, reaching 0.863 in the overall consideration. These results have shown that such workflow recognition obtained using this type of spatio-temporal instrument-aware information can provide better accuracy and nice clinical interpretation, and that it is well embedded in intelligent computer-assisted surgery. \\

Despite the successful outcomes provided by the proposed framework, it was tested on only one benchmark data set, and lack of time interval flexibility could impede long-distance procedural modeling operations. In the future, multi-center clinical validation, adaptive temporal modeling, domain adaptation, and self-supervised learning will be explored to further enhance the system's robustness and generalization. More integration of surgical work flow recognition and localization will also be written about so as to enhance reliable decision assistance in future robotic surgery system and the intelligent operating rooms.

\section*{Statements and Declarations}

\subsection*{Funding}
The authors declare that no funds, grants, or other support were received during the preparation of this manuscript.

\subsection*{Competing Interests}
The authors have no competing interest.

\subsection*{Author Contributions}
All authors contributed to the study conception and design. S.B. and M.S. performed coding implementation, material preparation, data collection, and analysis. The first draft of the manuscript was written by S.B. and M.S., while all authors commented on previous versions of the manuscript. All authors read and approved the final manuscript while S.B. supervised the entire research process.

\subsection*{Data Availability}
The Cholec80 dataset analyzed during the current study is publicly available. Requests for access can be directed to the dataset creators.

\subsection*{Code Availability}
The code will be made publicly available upon acceptance at \url{https://github.com/ac005sheekar/LapChol-Surgery-Swin-Cholec80}. The surgical phase recognition demonstration will be released at \url{https://youtu.be/ZyNXr5PbIHA}

\subsection*{Ethics Approval}
Not applicable. This study used a publicly available dataset (Cholec80) and did not involve human participants or animals.

\end{document}